\documentclass[preprint,5p,times,twocolumn]{elsarticle}

\usepackage{amssymb}
\usepackage{amsmath}
\usepackage{lineno}

\usepackage{listings}
\usepackage{multirow}
\usepackage{algorithmic}
\usepackage{algorithm}
\usepackage{url}

\lstdefinelanguage{ttl}{}
\journal{}

\begin{document}

\begin{frontmatter}



\title{Comparison of Metadata Representation Models for Knowledge Graph Embeddings}
% \title{Comparison of Knowledge Graph Representations for Link Prediction}
% \title{Efficient Knowledge Graph Representation for Link Prediction}


\author[aist]{Shusaku Egami} %% Author name
\author[fujitsu,aist]{Kyoumoto Matsushita}
\author[fujitsu,aist]{Takanori Ugai}
\author[aist]{Ken Fukuda}

\affiliation[aist]{organization={Artificial Intelligence Research Center(AIRC), National Institute of Advanced Industrial Science and Technology (AIST)},%Department and Organization
            addressline={2-4-7 Aomi}, 
            city={Koto},
            postcode={135-0064}, 
            state={Tokyo},
            country={Japan}}

\affiliation[fujitsu]{organization={Fujitsu Limited},%Department and Organization
            addressline={4-1-1 Kamikodanaka, Nakahara}, 
            city={Kawasaki},
            postcode={211-8588}, 
            state={Kanagawa},
            country={Japan}}

%% Abstract
\begin{abstract}

Hyper-relational Knowledge Graphs (HRKGs) extend traditional KGs beyond binary relations, enabling the representation of contextual, provenance, and temporal information in domains, such as historical events, sensor data, video content, and narratives.
HRKGs can be structured using several Metadata Representation Models (MRMs), including Reification (REF), Singleton Property (SGP), and RDF-star (RDR). However, the effects of different MRMs on KG Embedding (KGE) and Link Prediction (LP) models remain unclear.
This study evaluates MRMs in the context of LP tasks, identifies the limitations of existing evaluation frameworks, and introduces a new task that ensures fair comparisons across MRMs. Furthermore, we propose a framework that effectively reflects the knowledge representations of the three MRMs in latent space.
Experiments on two types of datasets reveal that REF performs well in simple HRKGs, whereas SGP is less effective. However, in complex HRKGs, the differences among MRMs in the LP tasks are minimal.
Our findings contribute to an optimal knowledge representation strategy for HRKGs in LP tasks. 

\end{abstract}


% \begin{highlights}
% \end{highlights}

%% Keywords

\end{frontmatter}




\section{Introduction}

A Knowledge Graph (KG) is ``a graph of data intended to accumulate and convey knowledge of the real world, whose nodes represent entities of interest and whose edges represent potentially different relations between these entities.''~\cite{hogan2021knowledge}
KGs enable the integration of heterogeneous data and representation of semantic relationships. KGs are essential for the development of intelligent systems as technology for structuring and managing large-scale knowledge. They serve as crucial foundational technologies for advancing artificial intelligence by enhancing knowledge retrieval, reasoning and explainability.

\begin{figure}[!t]
\centering
\includegraphics[width=\linewidth]{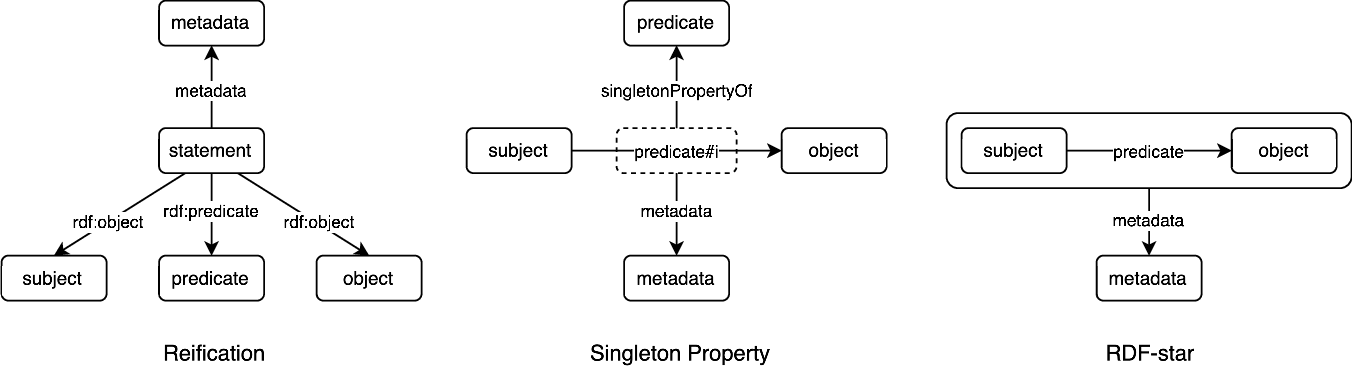}
\caption{Example of metadata representation models}
\label{fig:mrm}
\end{figure}

General KGs, such as DBpedia and Freebase, are typically structured as triples $\langle s, p, o \rangle$, consisting of a subject $s$, predicate $p$, and object $o$. 
However, representing knowledge using only simple triples is difficult in some cases. 
For example, it is often necessary to represent beyond the binary relations of triples in domains where context, provenance, and temporal information are required, such as historical events, sensor data, video content, and narratives.
In this paper, we collectively refer to KGs that represent such information as Hyper-Relational Knowledge Graphs (HRKGs).
In HRKGs, metadata such as provenance, context, and time must be associated with triples $\langle s, p, o \rangle$. 
Several metadata representation models (MRMs) have been proposed for this purpose, including RDF Reification\footnote{\url{https://www.w3.org/TR/rdf11-mt/\#reification}}, Singleton Property~\cite{nguyenDonRDFReification2014}, and RDF-star\footnote{\url{https://w3c.github.io/rdf-star/cg-spec/editors\_draft.html}}~\cite{hartig2017rdf}. 
Figure~\ref{fig:mrm} shows MRMs.
The choice of MRM in HRKG construction is a key consideration because it affects the data management and downstream tasks. 
Previous benchmarking studies have evaluated aspects such as query execution time, dataset size, and loading time into triple stores for different MRMs~\cite{freyEvaluationMetadataRepresentations2019,orlandiBenchmarkingRDFMetadata2021}, contributing to the appropriate choice of knowledge representation strategies for HRKGs.

On the other hand, research on KG embedding methods has rapidly advanced in recent years~\cite{wangKnowledgeGraphEmbedding2017a}, and the incorporation of KG embeddings (KGE) is expected to improve the performance of language models~\cite{zhang-etal-2019-ernie}.
In future artificial intelligence systems, KGs are expected to evolve beyond their traditional roles as knowledge bases and serve as modules that enhance the latent spaces of language models.
Thus, embedding KGs into latent spaces is an important research topic.

Although existing studies have evaluated various KG embedding models for link prediction, few evaluations have focused on MRMs.
Consequently, the optimal knowledge representation strategy for KG embeddings remains unclear.

To date, there has been no KGE model that can be applied to any MRM (especially RDF-star) ~\cite{10.1007/978-3-031-47240-4_15}, and there are issues of fairness in the evaluation that we point out in this study; therefore, a comprehensive evaluation is still challenging.
When embedding knowledge graphs into latent spaces, it is essential not only to improve the KGE models themselves but also to understand how the knowledge representation method affects the performance of downstream tasks.

To address this issue, this study constructs multiple datasets by applying different MRMs to HRKGs and evaluates the accuracy of link prediction (LP) using our proposed KGE method, which can be applied to any MRM, and then analyzes the characteristics of each MRM.
Specifically, we focus on RDF Reification, the Singleton Property, and the RDF-star model, as in the previous benchmarking~\cite{9364401}.
We conduct experiments using existing HRKG embedding models to identify the problems of typical LP experiments in MRM comparisons (we call these experiments Evaluation Method Diagnosis: EMD) and define a new experimental task that considers fairness in the evaluation.
Furthermore, we propose a framework containing KGE and LP models applicable to all three MRMs and conduct evaluation experiments under the defined task to clarify the characteristics of each MRM.
In particular, our study provides new insights into RDF-star embeddings, which have not been thoroughly evaluated in previous studies.
Our experimental code is available on GitHub\footnote{\url{https://github.com/aistairc/MRM-experiments}}.

The contributions of this study are summarized as follows:

\begin{itemize}
    \item Through the EMD, we identified issues related to tasks, embedding models, and datasets when applying conventional link prediction methods to knowledge graphs constructed with different MRMs.
    \item We constructed a complex HRKG dataset and designed a fair evaluation task across different MRMs.
    \item We developed a framework containing KGE and LP models applicable to all MRMs.
    \item We conducted evaluation experiments and analyzed datasets and hyperparameters to reveal the characteristics of each MRM.
\end{itemize}

The remainder of this paper is structured as follows.  
Section~\ref{sec:related_works} introduces MRMs and related works. 
Section~\ref{sec:prelimnary} describes EMD to identify the limitations of existing evaluation setups.  
Section~\ref{sec:approach} introduces the link prediction task and presents the proposed framework along with the experimental results.  
Section~\ref{sec:discussion} analyzes and discusses the experimental results.  
Finally, Section~\ref{sec:conclusion} concludes the paper, including a brief summary and future works.

\section{Related Work}
\label{sec:related_works}

\subsection{Metadata Representation Model}
\label{sec:mrm}

This section introduces MRMs, which are the subject of this study.

\subsubsection{Reification}

RDF Reification ( REF) is an MRM natively supported by RDF, where the subject, predicate, and object of a triple are reified to describe additional information. For example, when providing metadata $v$, such as provenance, for a standard triple $\langle s, p, o \rangle$, REF allows the following representation in Turtle\footnote{\url{https://www.w3.org/TR/turtle/}} syntax.

\begin{lstlisting}[caption=Example of Turtle notation for RDF Reification model,label=lst:REF]
<statement> rdf:type rdf:Statement ;
    rdf:subject <s> ;
    rdf:predicate <p> ;
    rdf:object <o> ;
    <metadata> <v> .
\end{lstlisting}

Because REF utilizes the standard RDF notation, the reified data can be stored in any RDF store.

\subsubsection{Singleton Property}

REF has the disadvantages of increasing the number of triples and lengthening the triple patterns in SPARQL queries. 
To address this issue, the Singleton Property~\cite{nguyenDonRDFReification2014} (referred to as SGP) was proposed as an MRM.
In the SGP, the example in Listing~\ref{lst:REF} is represented as follows:

\begin{lstlisting}[caption=Example of Turtle notation for Singleton Property model]
<s> <p#1> <o> .
<p#1> <singletonPropertyOf> <p> ;
    <metadata> <v> .
\end{lstlisting}

$p\#1$ is referred to as a singleton property, where the property $p$ is instantiated for each individual triple.  
A singleton property is assigned a unique Uniform Resource Identifier (URI) that allows metadata to be associated with it.

\subsubsection{RDF-star}

RDF-star (formerly known as RDF*)~\cite{hartig2017rdf} model has been proposed as an extension of RDF, allowing the representation of triples of triples using the Turtle-star\footnote{\url{https://w3c.github.io/rdf-star/cg-spec/editors_draft.html#turtle-star}} notation, as shown below.

\begin{lstlisting}[caption=Example of Turtle-star notation for RDF-star model]
<< <s> <p> <o> >> <metadata> <v> .
\end{lstlisting}

Here, a triple enclosed in $<< >>$ is referred to as a Quoted Triple (QT)\footnote{The quoted triples are sometimes referred to as embedded triples; however, to avoid confusion with knowledge graph embeddings, we use the abbreviation QT throughout this paper. In addition, although it is called a Triple Term in the latest draft, we use the previous name to avoid confusion in this paper.} and the metadata can be associated with QT.
In recent years, RDF-star has attracted significant attention because the RDF-star Working Group\footnote{\url{https://www.w3.org/2022/08/rdf-star-wg-charter/}} was established in August 2022 and is working on extending RDF as RDF 1.2.
RDF-star is still in the draft stage for W3C recommendation, and discussions on its specification are ongoing; however, it has already been implemented in many triplestores\footnote{\url{https://w3c.github.io/rdf-star/implementations.html}}.

\subsection{Knowledge Graph Embedding Models}

Knowledge graph embedding (KGE) models have been developed to obtain the distributed representations of entities and relations in KGs.  
Several hyper-relational KGE models were proposed~\cite{NaLP, galkin-etal-2020-message}.  
NaLP~\cite{NaLP} is a KGE model that explicitly models the relatedness of role-value pairs within the same hyper-relation. The NaLP employs a fully connected neural network.
StarE~\cite{galkin-etal-2020-message} was proposed as a link prediction method for HRKGs, such as Wikidata.  
StarE consists of an encoder that extends CompGCN~\cite{vashishth2020compositionbased} to handle {\it qualifiers} and a Transformer-based decoder.
These KGE models cannot handle complex HRKGs based on each MRM targeted in this study.  
Therefore, we propose KGE and LP models that are adaptable to HRKGs based on REF, SGP, and RDR.

\subsection{Benchmarking MRMs}

Frey et al.~\cite{freyEvaluationMetadataRepresentations2019} conducted benchmarking on various metadata representation models (MRMs), including REF, RDR, SGP, N-ary relations, and Named Graphs, focusing on the number of triples, dataset size, loading time into a triple store, and query execution time.  
They used edit history data of DBpedia and Wikidata as datasets and reported the results of comparative experiments applying different MRMs.

Orlandi et al.~\cite{9364401} applied REF, SGP, and RDR to the Biomedical Knowledge Repository dataset and benchmarked the dataset size and query execution time.  
These studies compared different MRMs with a focus on triple stores rather than KGE.

Iglesias-Molina et al.~\cite{10.1007/978-3-031-47240-4_15} analyzed REF, RDR, N-ary relations, and Wikidata qualifiers based on various user scenarios, including knowledge exploration, systematic querying, and graph completion.
However, a comprehensive analysis of graph completion was not conducted owing to the lack of a KGE model for the RDF-star.
In addition, fairness issues, as pointed out in Section \ref{sec:problem}, remain unaddressed.

\section{Evaluation Method Diagnosis}
\label{sec:prelimnary}

To identify the limitations of existing evaluation setups and to motivate the design of a fair comparison across MRMs, we conduct EMD.

In the EMD, we prepare KGs adopting REF, SGP, and RDR on the same dataset and evaluate the link prediction accuracy.  
Here, we conduct EMD using StarE on the same task as in previous studies~\cite{galkin-etal-2020-message} and describe the issues identified through these experiments.

\subsection{Task Definition}

In EMD, link prediction (LP) is performed using HRKG datasets converted into different MRMs.
As the dataset, we use WD50K~\cite{galkin-etal-2020-message}.  
WD50K is a dataset sampled from Wikidata for evaluating hyper-relational LP models, constructed to address the limitations of existing datasets: WikiPeople~\cite{NaLP} contains numerous literal values, and JF17K~\cite{10.5555/3060621.3060802} suffers from test leakage.
Wikidata contains {\it qualifiers} that qualify triples, allowing supplementary information such as timestamps to be associated with individual triples.  
Thus, Wikidata is considered to be an HRKG.  
WD50K is provided in the following CSV format:
\begin{lstlisting}[caption=Example data from WD50K, label=listing:wd50k]
Q1968853,P166,Q3703462,P1346,Q55245
\end{lstlisting}
Q1968853 (Richard III) is the subject ($s$) of the triple, P166 (award received) is the predicate ($p$), and Q3703462 (David di Donatello for Best Foreign Production) is the object ($o$).  
In addition, P1346 (winner) serves as the {\it qualifier} ($qr$), and Q55245 (Laurence Olivier) is its corresponding value ($qv$).

In this EMD, we follow the experimental setup of~\cite{galkin-etal-2020-message}, using $s, p, qr_1, qv_1, qr_2, qv_2, \dots, qr_n, qv_n$ as input and predicting $o$ as the output.
We used the Mean Reciprocal Rank (MRR) and Hits@N as evaluation metrics.  
The MRR ranks the predicted entities and computes the mean of the reciprocal ranks of the correct entity.  
A higher MRR score indicates that the correct entity is ranked higher.  
Hits@N measures the proportion of cases where the correct entity appears within the top $N$ predictions.
In this experiment, we set $N = \{1, 10\}$.

\subsection{Dataset Preparation}
\label{sec:wd50k_data}

In this study, we focus on REF, SGP, and RDR as MRMs and convert the CSV data of WD50K into RDF data for each MRM.
The conversion from WD50K CSV data to REF-based RDF data (Turtle syntax) can be expressed as follows:
\begin{equation}
    \begin{aligned}
        (s, p, o, qr_1, qv_1, qr_2, qv_2, \dots, qr_n, qv_n) \\
        \mapsto~st \; \text{rdf:subject} \; s \; ; \; 
        \text{rdf:predicate} \; p \; ; \; & \text{rdf:object} \; o \; ; \\
        \; qr_1 \; qv_1 \; ; \; qr_2 \; qv_2 \; ;& \dots ; qr_n \; qv_n \; .
    \end{aligned}
\end{equation}
where $st$ is a newly created statement instance as a blank node.
For example, converting the data in Listing~{listing:wd50k} into REF results in the following:
\begin{lstlisting}[caption=Example of REF-based WD50K's KG]
_::B2a396[...]bac53f 
    rdf:subject wd:Q1968853 ; 
    rdf:predicate wd:P166 ; 
    rdf:object wd:Q3703462 ; 
    wd:P1346 wd:Q55245 .
\end{lstlisting}

The conversion from the WD50K CSV data to SGP-based RDF data can be expressed as follows:
\begin{equation}
\begin{aligned}
    (s, p, o, qr_1, qv_1, qr_2, qv_2, \dots, qr_n, qv_n) \\
\mapsto
    s \; p\#k \; o \; . \;
    p\#k \; \text{singletonPropertyOf}& \; p \; ;\\
    \; qr_1 \; qv_1 \; ; \; qr_2 \; qv_2 \; ; \dots ; qr_n \; qv_n \; .
\end{aligned}
\end{equation}
where k is an identifier used to specialize p for each triple, and an integer value is used.
For example, converting the data in Listing~\ref{listing:wd50k} into the SGP results in the following.

\begin{lstlisting}[caption=Example of SGP-based WD50K's KG]
wd:Q1968853 wd:P166#1 wd:Q3703462 .
wd:P166#1 :singletonPropertyOf wd:P166 ;
    wd:P1346 wd:Q55245 .
\end{lstlisting}

The conversion from WD50K CSV data to RDR data can be expressed as follows.
\begin{equation}
    \begin{aligned}
        (s, p, o, qr_1, qv_1, qr_2, qv_2, \dots, qr_n, qv_n) \\
        \mapsto~\ll s \; p \; o \gg \; qr_1 \; qv_1 \; ; \; qr_2 \; qv_2 \; ; \dots &; qr_n \; qv_n \; .
    \end{aligned}
\end{equation}
For example, converting the data in Listing~\ref{listing:wd50k} into the RDR results in the following.
\begin{lstlisting}[caption=Example of RDR-based WD50K's KG]
<< wd:Q1968853 wd:P166 wd:Q3703462 >> 
    wd:P1346 wd:Q55245 .
\end{lstlisting}

Finally, we convert each MRM-based KG back to the original CSV format for use as an input for StarE.
Consequently, the CSV data for each MRM is formatted as shown in Listings~\ref{lst:ref-stare-csv}--\ref{lst:rdr-stare-csv}.  
Note that the CSV data for RDR follows the same ordering as the original WD50K CSV.

\begin{lstlisting}[caption=Example of CSV data for REF-based WD50K,label=lst:ref-stare-csv]
_:B2a396[...]bac53f,rdf:object,wd:Q3703462,rdf:subject,wd:Q1968853,rdf:predicate,wd:P166,wd:P1346,wd:Q55245
\end{lstlisting}

\begin{lstlisting}[caption=Example of CSV data for SGP-based WD50K,label=lst:sgp-stare-csv]
wd:Q1968853,wd:P166#1,wd:Q3703462,singletonPropertyOf,wd:P166,wd:P1346,wd:Q55245
\end{lstlisting}

\begin{lstlisting}[caption=Example of CSV data for RDR-based WD50K,label=lst:rdr-stare-csv]
wd:Q1968853,wd:P166,wd:Q3703462,wd:P1346,wd:Q55245
\end{lstlisting}

\subsection{Result}
\label{sec:preliminary_result}

For StarE, we use the default hyperparameter settings.  
The optimization algorithm is Adam~\cite{kingma2014adam}, the loss function is binary cross-entropy loss, and the number of epochs is set to 400.  
The experiments were conducted using four NVIDIA V100 GPUs.

Table~\ref{table:stare} lists the link prediction results using StarE for each MRM.
The results indicate that in WD50K, RDR (i.e., the original data format of WD50K) provides a structure that is well suited for predicting the object entity $o$.
Conversely, REF showed a significant decrease in the MRR and Hits@N, suggesting that it is an unsuitable structure for predicting $o$.
One possible reason for this result is that, in REF for WD50K, individual statement entities do not appear in other triples.
Consequently, StarE cannot sufficiently learn their meanings, leading to a substantial decrease in accuracy.

\begin{table}[t]
\caption{Preliminary experimental results using StarE}
\begin{center}
\begin{tabular}{llll}
\hline
    & MRR    & Hits@1 & Hits@10  \\ \hline
REF & 0.11 & 0.066 & 0.20   \\
SGP & 0.58 & 0.51 & 0.72   \\
RDR & 0.65  & 0.58  & 0.77   \\ \hline
\end{tabular}
\end{center}
\label{table:stare}
\end{table}
 %Preliminary experimental results using StarE

\subsection{Problem}
\label{sec:problem}

The results of EMD revealed issues related to tasks, KGE models, and datasets.

\subsubsection{Task consistency and fairness}

In the EMD, we formulated the task as predicting the same object entity $o$ across three MRMs.
However, the assumption of task equivalence raises concerns when considering the semantics of target triples.
Specifically, in the case of RDR (Listing~\ref{lst:rdr-stare-csv}), the semantics of the triple containing the object entity $o$ is  
``Richard III, award received, David di Donatello for Best Foreign Production.''  
In contrast, in the case of REF (Listing~\ref{lst:ref-stare-csv}), the semantics of the triple is  
``\_:B2a396[...]bac53f, object, David di Donatello for Best Foreign Production.'' 
Therefore, the semantics and contextual interpretation of the target triples vary significantly across different MRMs, making it difficult to ensure that they are evaluated under the same task.

Furthermore, as mentioned in Section~\ref{sec:preliminary_result}, each statement entity in the REF data appears only once as the subject of a triple.
Consequently, the KGE model lacks sufficient opportunities to learn meaningful representations of statement entities.  
In contrast, in the RDR data, the subject entity $s$ appears in other triples, providing adequate training data for effective learning.  
This discrepancy in the availability of training samples across different MRMs challenges the fairness of evaluation.

\subsubsection{Complexity of dataset}

Although WD50K contains metadata in the form of $qr$ and $qv$ associated with triples, it has a simple structure, where $qv$ does not appear in the other triples' $s$ and $o$.
In other words, entities such as statement, singleton property, and QT (hereafter referred to as triple entities: TEs) never appear as $qv$ in WD50K. 
This means that no structures corresponding to TE--TE relationships exist.
Furthermore, TEs do not appear as $o$, and nested triple structures are not present.
Note that RDF-star can represent nested structures because QT can be used as compositional entities of another QT.
Since such structures can occur in real-world data, it is necessary to conduct evaluation experiments using more complex datasets.

\subsubsection{Limitations of KGE models}

In the CSV data format used as input for StarE (Listing~\ref{lst:rdr-stare-csv}), QT is not represented as entities.  
As a result, the original RDF-star knowledge representation is compromised.  
Therefore, in EMD using StarE, it is challenging to obtain embedding vectors that accurately reflect the original knowledge representation.
In addition, StarE cannot handle datasets in which QTs appear as $o$ or $qv$.  
Furthermore, in the SGP model, the knowledge representation in which singleton properties are used not only as relations but also as entities is missing when converting the data into StarE's input format.

\section{Approach}
\label{sec:approach}

\begin{figure*}[!t]
\centering
\includegraphics[width=\linewidth, trim=10 610 10 20, clip]{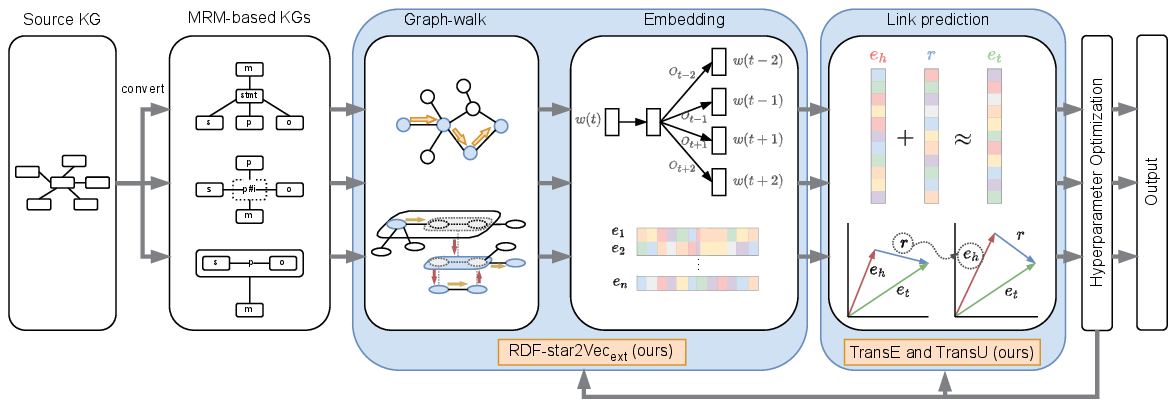}
\caption{Overview of experimental approach}
\label{fig:approach}
\end{figure*}

In this study, to address the challenges identified in Section~\ref{sec:problem} and explore the characteristics of each MRM in detail, (1) we construct a more complex HRKG dataset based on each MRM, (2) design a link prediction task that ensures that the target entities are identical and the evaluation is fair across different MRMs, and (3) conduct experiments using our KGE model capable of appropriately reflecting the knowledge representations of all MRMs in the latent space.  
Figure~\ref{fig:approach} shows an overview of our experimental approach.

\subsection{Task Definition}

The task in this experiment is a standard link prediction task, where $s$ and $p$ are given as input to predict $o$.  
However, in this task, the target triples are adjusted to ensure that they have the same semantics across different MRMs.  
The target triples $\langle s, p, o \rangle$ for each MRM are as follows:

\begin{itemize}
    \item REF
        \begin{itemize}
            \item [$s$:] $e \in \mathcal{E}$
            \item [$p$:] $r \in \mathcal{R} \setminus \{\text{rdf:subject}, \text{rdf:predicate}, \text{rdf:object}\}$
            \item [$o$:] $e \in \mathcal{E}$
        \end{itemize}
    \item SGP
        \begin{itemize}
            \item [$s$:] $e \in \mathcal{E}$
            \item [$p$:] $r \in \mathcal{R} \setminus (\{\text{:singletonPropertyOf}\} \cup \mathcal{R}_{\text{SGP}})$\\ 
            where $\mathcal{R}_{\text{SGP}} = \{ p_{\#i} \mid \exists s, o, p \in \mathcal{R}, (s, p_{\#i}, o) \in \mathcal{T} \text{ and } (p_{\#i}, \text{:singletonPropertyOf}, p) \in \mathcal{T} \}$
            \item [$o$:] $e \in \mathcal{E}$
        \end{itemize}
    \item RDR
        \begin{itemize}
            \item [$s$:] $e \in \mathcal{E}$
            \item [$p$:] $r \in \mathcal{R} \setminus \mathcal{R}_{\text{QT}}$\\ where $\mathcal{R}_{\text{QT}} = \{ p \mid \exists s, o, qr, qv, (( s, p, o ), qr, qv) \in \mathcal{T} \}$
            \item [$o$:] $e \in \mathcal{E}$
        \end{itemize}
\end{itemize}
, where $\mathcal{E}$ denotes the set of entities, $\mathcal{R}$ denotes the set of relations, and $\mathcal{T}$ denotes the set of triples.
$\mathcal{R}_{SGP}$ denotes the set of singleton properties and $\mathcal{R}_{QT}$ denotes the set of relations in which the subject is a QT.
This approach enables a fair experimental setup, ensuring that the task remains consistent across different MRMs. 
For WD50K, we use WD50K(100), in which all triples are hyper-relational.  
In the KGRC-RDF, some standard triples may be included; however, these triples appear identically across all MRMs.  
Thus, the primary prediction targets are the metadata values associated with the triples.

\subsection{Data Preparation}

We use WD50K, which was employed in the EMD, and KGRC-RDF~\cite{kawamura2022contextualized}, which is a more complex HRKG dataset.

\subsubsection{WD50K}

The KGs generated by adopting each MRM to WD50K, as described in Section~\ref{sec:wd50k_data}, are formatted into CSV data containing $\langle s, p, o \rangle$ triples.  
Here, the dataset is split to ensure that each TE as $s$ is contained in both the training and test sets.  
This ensures that the amount of training data for $s$ remains balanced across different MRMs.

\subsubsection{KGRC-RDF}
KGRC-RDF is a type of Event KG~\cite{guanWhatEventKnowledge2022} based on the content of mystery stories and is provided in the following RDF format.

\begin{lstlisting}[caption=Example from KGRC-RDF, label=lst:kgrc-rdf]
@prefix kdrp: <http://kgc.knowledge-graph.jp/data/ResidentPatient/> .
@prefix kdp: <http://kgc.knowledge-graph.jp/data/predicate/> .
@prefix kgc: <http://kgc.knowledge-graph.jp/ontology/kgc.owl#> . 
kdrp:105 
  kgc:source "The young man was caring for an elderly man"@en;
  rdf:type kgc:Situation ;
  kgc:hasPredicate kdp:care ;
  kgc:subject kdp:Young_man ;
  kgc:then kdrp:106 ;
  kgc:what kdrp:Elderly_man .
\end{lstlisting}
The representation of KGRC-RDF is equivalent to REF and contains detailed descriptions of ``who, what, when, where, why, and how (5W1H)'' information along with the storyline and has a rich variation of entities as values of 5W1H.
KGRC-RDF is a benchmark dataset for Explainable AI (XAI) development, designed for a competition~\cite{kozaki2023datasets} in which participants predict the culprit using an incomplete KG and logically explain the reasoning behind the prediction.

The original KGRC-RDF dataset is directly used as the REF model, and it is further converted into SGP and RDF-star models.
In the SGP and RDR models, it is necessary to explicitly define the object entity $o$.
In this study, the priority order for object selection is set as $what > whom > where > on > to > from$.
For example, the conversion result from Listing~\ref{lst:kgrc-rdf} into SGP is as follows:
\begin{lstlisting}[caption=Example of SGP-based KGRC-RDF's KG]
kdrp:Young_man kdrp:care-1 kdrp:Eldery_man .
kdrp:care-1 :singletonPropertyOf kdrp:care ;
  rdf:type kgc:Situation ;
  kgc:source "The young man was caring for an elderly man"@en ;
  kgc:then kdrp:say-3 ;
\end{lstlisting}

Similarly, the conversion result from Listing~\ref{lst:kgrc-rdf} into RDR is as follows:  
The representation of the RDR data of KGRC-RDF is the same as that of the existing KGRC-RDF-star dataset ~\cite{rdf-star2vec}.

\begin{lstlisting}[caption=Example of RDR-based KGRC-RDF's KG]
<< 
  << kdrp:Young_man kdp:care kdrp:Elderly_man >> rdf:value "http://kgc.knowledge-graph.jp/data/ResidentPatient/105" 
>> rdf:type kgc:Situation ;
  kgc:source "The young man was caring for an elderly man"@en ;
  kgc:then 
    << 
      << kdrp:Young_man kdp:say << 
        << kdrp:Elderly_man kgc:hasProperty kdp:equalTo >> rdf:value "http://kgc.knowledge-graph.jp/data/ResidentPatient/107" >>
      >> rdf:value "http://kgc.knowledge-graph.jp/data/ResidentPatient/106" 
    >>
\end{lstlisting}
RDR allows new triples to be expressed using QT in the form of $\langle \verb|<<|s ~ p ~ o\verb|>>| ~ qr ~ qv\rangle$.  
A QT is uniquely determined by the combination of $s, p, o$, and no URI is assigned to represent a QT.  
However, when converting KGRC-RDF into RDR, there are cases where the same combination of $s, p, o$ appears in different contexts.
To ensure an accurate semantic representation, it is necessary to distinguish between these QTs and assign different metadata to them.  
To address this issue, KGRC-RDF-star introduces a unique identifier to each QT, structuring the triples in a nested format as $\langle\verb|<<| ~ \verb|<<|s ~ p ~ o\verb|>>| ~ id ~ val \verb|>>| ~ qr ~ qv\rangle$.

Similar to WD50K, the RDR dataset is split to ensure that each QT appearing as a subject entity $s$ is present in both the training and test sets.

\subsection{Proposed Framework}

We propose a novel LP framework that is applicable to REF, SGP, and RDR models.
In the proposed framework, we employ an extended version of RDF-star2Vec~\cite{rdf-star2vec} to pre-train the distributed representations of KGs and perform additional training using a translation-based model to adapt it to the LP task.
Each component of the framework is described in the following section.

\begin{table*}[t!]
\caption{Experimental results using our framework}
\label{table:transx}
\begin{center}
\begin{tabular}{rrrrrrrrr}
\hline
&& \multicolumn{3}{r}{WD50K} && \multicolumn{3}{r}{KGRC-RDF}  \\ \cline{3-5} \cline{7-9} 
&& MRR   & Hits@1& Hits@10   && MRR   & Hits@1& Hits@10   \\ \hline
REF & TransE & \textbf{0.53} & \textbf{0.42} & \textbf{0.75} && 0.32  & 0.21  & \textbf{0.59} \\
& TransU & 0.52  & 0.41  & 0.73  && 0.32  & 0.22  & 0.58  \\
SGP & TransE & 0.18  & 0.11  & 0.32  &  & \textbf{0.34} & \textbf{0.25} & 0.56  \\
& TransU & 0.13  & 0.08  & 0.24  && 0.33  & 0.24  & 0.52  \\
RDR & TransE & 0.46  & 0.33  & 0.71  && 0.31  & 0.23  & 0.52  \\
& TransU & 0.46  & 0.33  & 0.71  && 0.33  & 0.24  & 0.54  \\ \hline
\end{tabular}
\end{center}
\end{table*} %Experimental results using our framework

\subsubsection{Pre-training step}
In this step, we generate distributed representations of all entities and relations, including TE, for each MRM.

RDF-star2Vec is a KGE model capable of handling complex RDF-star graphs, where $s$, $o$, and $qv$ can be QTs while learning their embeddings as nodes.
Therefore, it enables the acquisition of embedding representations for complex knowledge graphs such as KGRC-RDF-star. 
RDF-star2Vec extends RDF2Vec~\cite{ristoskiRDF2VecRDFGraph2016}, a neural network-based KGE model, by introducing the following two graph walk strategies.

\begin{itemize}
    \item qs-walk: walk from a QT to its compositional entity in the subject role
    \item oq-walk: walk from a compositional entity in the object role to the QT
\end{itemize}

In this study, we further propose two additional graph walk strategies and introduce additional configurable parameters to explore the optimal embedding representations of RDR.
The RDF-star2Vec extended in this study is denoted as RDF-star2Vec$_{\text{ext}}$.

\begin{itemize}
    \item qo-walk: walk from a QT to its compositional entity in the object role
    \item os-walk: walk from a compositional entity in the subject role to the QT
\end{itemize}
The algorithms are described in \ref{appendix:walk}.
RDF-star2Vec$_{\text{ext}}$ inputs the generated set of graph walk sequences into structured word2vec~\cite{ling-etal-2015-two} which considers the word order and obtains the embedding representations of each entity, relation, and QT.

Let $\alpha$, $\beta$, $\gamma$, and $\delta$ denote the transition probabilities of qs-walk, oq-walk, qo-walk, and os-walk, respectively.
Applying RDF-star2Vec$_{\text{ext}}$ to MRM-based KGs other than the RDR sets these probabilities to zero, rendering the resulting embeddings equivalent to those of RDF2Vec.

\subsubsection{Link prediction step}

Neural network-based unsupervised graph representation learning models, such as RDF2Vec, are unsuitable for direct application to LP.  
Thus, we perform additional training on LP models using the embedding vectors pre-trained by RDF-star2Vec$_{\text{ext}}$.
In this experiment, we employ TransE~\cite{bordesTranslatingEmbeddingsModeling2013} as an existing link prediction model.  
Additionally, we developed a new LP model to leverage the knowledge representation specific to the SGP model.

In SGP, singleton properties are used both as entities and relations.  
In the implementation of TransE, the same singleton property is assigned different embedding vectors depending on whether it is treated as an entity or a relation.
To address this issue, we developed a method that assigns a unique embedding vector to entities that are also used as relations, referred to as TransU.

\subsubsection{Hyperparameter optimization}
\label{sec:optimization}

The proposed framework integrates multiple models and requires the configuration of several hyperparameters.
To compare the best results across different MRMs, we perform hyperparameter optimization.

The hyperparameters in RDF-star2Vec$_{\text{ext}}$ are set as follows:  
transition probability $\alpha,\beta,\gamma,\delta=[0,1]$, the number of walks $n \in \{10, 100, 200\}$, depth of the walks $d=[3,12]$, and the walk generation modes $mode \in \{$
STAR\_MID\_WALKS, 
STAR\_MID\_WALKS\allowbreak\_DUPLICATE\allowbreak\_FREE, \allowbreak
STAR\allowbreak\_RANDOM\_WALKS, \allowbreak 
STAR\allowbreak\_RANDOM\allowbreak\_WALKS\_DUPLICATE\allowbreak\_FREE 
$\}$.
For structured word2vec, the hyperparameters are set as follows:  
the dimension size $dim \in \{50,100,200,400\}$, the window size $w \in \{5,7,9,11\}$, and the neural network algorithm $l \in \{$CBOW, skip-gram, cwindow, structured skip-gram$\}$.
To optimize these hyperparameters, we employ Optuna and apply the Tree-Structured Parzen Estimator (TPE)~\cite{watanabe2023tree}.

\subsection{Result}
\label{sec:result}

Table~\ref{table:transx} shows the LP results obtained using our framework for each MRM.
In WD50K, applying TransE to REF achieved the highest accuracy, while SGP showed significantly lower accuracy compared to the other two MRMs.
In contrast, there was no significant difference between the MRMs in KGRC-RDF and WD50K.
RDR was less affected by variations in LP models and datasets, and showed overall stable results. 

\begin{table*}[t]
\caption{Statistics of datasets}
\label{table:statistics_of_data}
\begin{center}
\begin{tabular}{lrrrrr}
\hline
& \multicolumn{2}{c}{WD50K}& \multicolumn{1}{c}{} & \multicolumn{2}{c}{KGRC-RDF} \\ \cline{2-3} \cline{5-6} 
& \multicolumn{1}{c}{\# of entities} & \multicolumn{1}{c}{\# of relations} & \multicolumn{1}{c}{} & \multicolumn{1}{c}{\# of entities} & \multicolumn{1}{c}{\# of relations} \\ \hline
Reification   & 48,018 & 138 &  & 7,041  & 44  \\
SingletonProperty & 48,018 & 29,158  &  & 8,463  & 2,323   \\
RDF-star  & 47,814 & 279 &  & 7,449  & 575 \\ \hline
\end{tabular}
\end{center}
\end{table*} %Statistics of datasets
\begin{table}[t]
\caption{Proportion of triples containing QT (\%)}
\label{table:qt}
\begin{center}
\begin{tabular}{crrr}
\hline
\multicolumn{1}{l}{} & \multicolumn{1}{c}{train} & \multicolumn{1}{c}{valid} & \multicolumn{1}{c}{test} \\ \hline
QT $\rightarrow$ QT  & 17.51 & 28.32 & 27.09\\
QT $\rightarrow$ AT  & 28.93 & 47.24 & 51.08\\
AT $\rightarrow$ QT  & 6.03  & 0.15  & 0.15 \\
AT $\rightarrow$ AT  & 47.53 & 24.29 & 21.67\\ \hline
\end{tabular}
\end{center}
\end{table} %Proportion of triples containing QT (\%)

\begin{figure*}[t]
\begin{center}
\includegraphics[width=\linewidth]{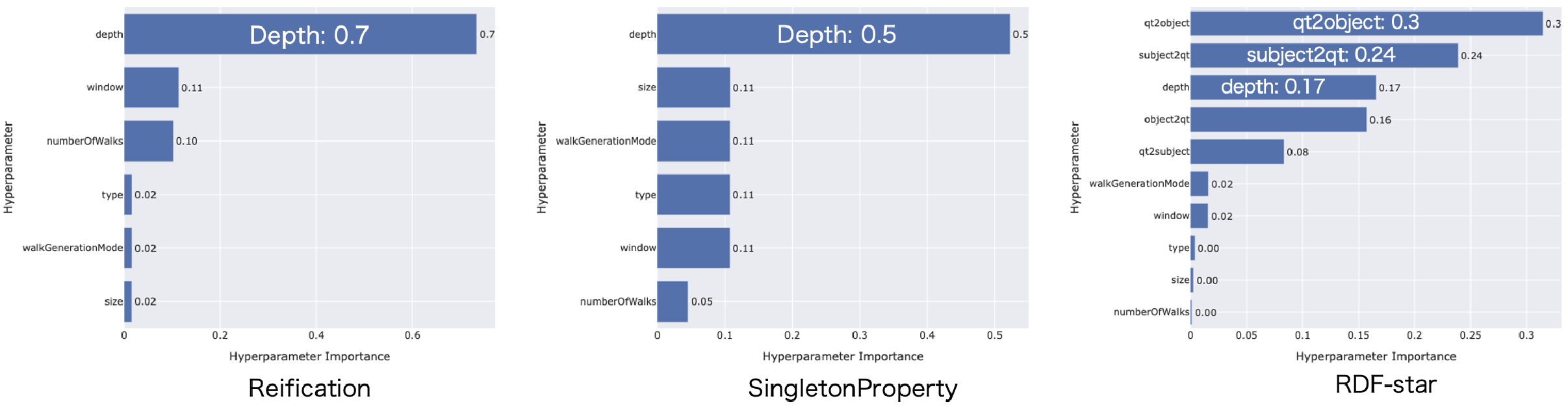}
\caption{Hyperparameter importance}
\label{fig:importance}
\end{center}
\end{figure*}

\section{Analysis and Discussion}
\label{sec:discussion}

\subsection{Analysis of datasets and LP models}
\label{sec:dataset_analysis}

Table~\ref{table:statistics_of_data} shows the number of entities and relations in each dataset.  
In the SGP data of WD50K, the number of relations increases significantly.  
Since SGP model instantiates relations for each triple, each singleton property is typically used only once as a relation.
Therefore, in the SGP model, both the KGE and LP models fail to sufficiently learn the semantics of singleton properties as relations, which is considered to lead to decreased accuracy.

TransU is an extension of TransE proposed to reflect the knowledge representation of SGP-based KGs.
However, in the SGP data of WD50K, TransU resulted in a lower accuracy than TransE.
Typically, entities and relations have different roles in the KG.  
Since TransE assumes $s + p \approx o$, the relation $p$ represents a vector transformation between entities.
Unifying the embeddings of the same singleton property for both entities and relations may have disrupted its function as a vector transformation.
In particular, as mentioned earlier, each singleton property appears only once as a relation.
Thus, the LP models have limited opportunities to learn singleton properties as relations.  
In contrast, TransE maintains distinct vectors for entities and relations.  
That is, after the pre-training step with RDF-star2Vec$_{\text{ext}}$, two types of embeddings are generated: one for singleton properties that continue to be trained as entities and another for those that are rarely updated as relations.
It is considered that the slight improvement in the accuracy of TransE over TransU is due to the separation of roles within the embeddings of singleton properties.

In the REF and RDR models, there is almost no difference in the TransE and TransU scores. This is because they contain almost no relations that are used as entities, such as singleton properties.

In KGRC-RDF, the overall accuracy decreased compared that with of WD50K. 
Additionally, no significant differences were observed among different MRMs.
Table~\ref{table:qt} shows the proportion of triples in the RDR data of KGRC-RDF.
AT means the entity of an asserted triple (i.e., standard triple $\langle s,p,o \rangle$) in the RDR-based KG.
While WD50K contains only QT $\rightarrow$ AT triples, KGRC-RDF includes three types of QT-containing triples.
In KGRC-RDF, the distribution of triples differs across the train, validation, and test sets, which may have affected the overall decrease in accuracy.

Compared to WD50K, there was not much difference between MRMs in KGRC-RDF.
Therefore, SGP remains a potential candidate as an MRM strategy for LP on complex KGs.

\subsection{Analysis of Hyperparameters}
\label{sec:hyperparameter}

Figure~\ref{fig:importance} shows the hyperparameter importance of the proposed framework across MRMs in KGRC-RDF.
The depth of the graph walk was found to be the most important hyperparameter for REF and SGP.
The LP accuracy tends to be highest at a depth of approximately 9 in REF, approximately 3 in SGP, and approximately 6 in RDR (see Figure~\ref{fig:depth_ref}–\ref{fig:depth_rdr}).
Thus, the optimal hyperparameter values for LP vary significantly depending on the MRM of the KG.

\begin{figure}[t]
\begin{center}
\includegraphics[width=\linewidth]{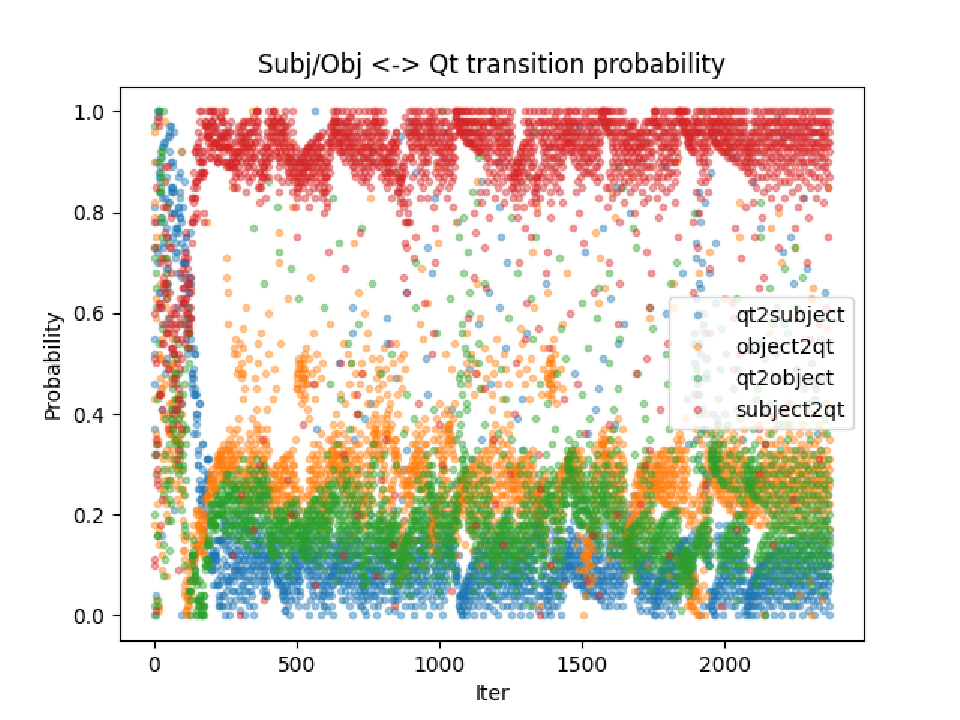}
\caption{Optimization process of walking strategy parameters}
\label{fig:qtwalk}
\end{center}
\end{figure}

Furthermore, in RDR, the newly introduced walking strategies in RDF-star2Vec$_{\text{ext}}$, qt2object and subject2qt, are more important than depth.
Figure~\ref{fig:qtwalk} shows the optimization process and the values of each walking strategy parameter in the RDR data of KGRC-RDF.
The LP accuracy tended to be highest when qt2object was approximately 0.14 (for more details, see Figure~\ref{fig:qt2object}).
In the walk sequences generated by qt2object, the QT entity and the triples composing that QT are not placed in close proximity.  
For example, if a QT $\verb|<<|e_{1}\ r_{1}\ e_{2}\verb|>>|$ exists, then qt2object walks follows $\verb|<<|e_{1}\ r_{1}\ e_{2}\verb|>>| \xrightarrow{} e_{2}$.  
Subsequently, this walk strategy transitions from $e_{2}$ to a relation unrelated to the QT.
Consequently, in the walk sequence, $e_{1}$ and $r_{1}$ are less likely to appear near $\verb|<<|e_{1}\ r_{1}\ e_{2}\verb|>>|$, potentially hindering the adequate learning of QT.
Setting this transition probability to a low value is a rational choice for improving performance.
On the other hand, subject2qt tended to achieve the highest link prediction accuracy when set at approximately 0.95 (for more details, as shown in Figure~\ref{fig:subject2qt}).
Even in the sequences generated by this walk, the QT entity and the triples composing that QT are not placed in close proximity.
For instance, when walking as $e_{1} \rightarrow r_{1} \rightarrow e_{2}$, if a QT $\verb|<<|e_{2}\ r_{2}\ e_{3}\verb|>>|$ exists, then the transition by subject2qt follows $e_{2} \rightarrow \verb|<<|e_{2}\ r_{2}\ e_{3}\verb|>>|$.
However, an unexpected result was observed: higher subject2qt probabilities correlated with improved LP accuracy.
Since the primary prediction targets in this task are metadata entities, the increased frequency of transitions to QTs and subsequent walks on metadata entities is considered to have directly contributed to the accuracy improvement.

Despite transitioning to QT in the same way as subject2qt, the object2qt parameter tended to achieve the highest LP accuracy when set at approximately 0.27 (for more details, see Figure~\ref{fig:subject2qt}).
In KGRC-RDF, for deeply nested QTs, the entity $e_{2}$ in $\verb|<<|e_{1}\ r_{1}\ e_{2}\verb|>>|$ is often itself a QT.\footnote{Here, double-nested structures where a URI is assigned for QT identification are excluded.}  
When the walk follows $e_{1} \rightarrow r_{1} \rightarrow e_{2} \rightarrow \verb|<<|e_{1}\ r_{1}\ e_{2}\verb|>>|$ by object2qt, if $e_{2}$ is a QT, QTs will appear consecutively in the walk sequence.
On the other hand, in subject2qt, QTs rarely appear consecutively in the walk sequence.  
This difference may have influenced the optimal values of the respective hyperparameters.

\section{Conclusion}
\label{sec:conclusion}

In this study, we focused on three MRMs applicable to HRKGs—REF, SGP, and RDR—and compared them from the perspective of KGE to clarify their differences in characteristics.
In the EMD, we identified issues in existing link prediction tasks and proposed a new task.  
Furthermore, we proposed a KGE framework capable of reflecting the unique knowledge representation of each MRM in the latent space.  
We analyzed and discussed the results based on datasets and hyperparameters.  
A concise summary of key findings from our experimental results is as follows: 
\begin{itemize}
\item In HRKGs with a simple structure, the knowledge representation of REF is suitable for the LP task, whereas SGP is not.  
\item In HRKGs with a complex structure, there is not much difference among the MRMs in the LP task.
\end{itemize}

To compare the LP accuracy across different MRMs, we conducted experiments using the KGE model, which was as easy to interpret as possible.
In future work, we plan to explore experiments using more advanced models, particularly LLM-based LP models~\cite{shu2024knowledgegraphlargelanguage,10884231}, and investigate whether fair evaluation is possible across all MRMs.

\section*{Acknowledgments}
This paper is based on results obtained from a project, JPNP20006, commissioned by the New Energy and Industrial Technology Development Organization (NEDO), and JSPS KAKENHI Grant Numbers JP22K18008 and JP23H03688.

\appendix
\section{Algorithm of graph walk added in RDF-star2Vec$_{\text{ext}}$}
\label{appendix:walk}

In Algorithm~\ref{algorithm:gt_walk_generation}, the input is as follows: a list of walk sequences for a root node is $wl$, a walk sequence is $walk$, a list of triples which involving $e$ as a subject is $triples_{e}^{subj}$, a QT which involving $e$ as an object is $qt_{e}^{obj}$, a QT which involving $e$ as a subject is $qt_{e}^{subj}$, and a transition probability of qs-walk is $\alpha$, a transition probability of qs-walk is $\beta$, a transition probability of qo-walk is $\gamma$, and a transition probability of sq-walk is $\delta$.
Where $qt_{e}^{subj}$ means a QT that involves $e$ as a subject, $triples_{e}^{subj}$ means a list of triples that involve $e$ as a subject, $qt_{e}^{obj}$.qt means $qt_{e}^{obj}$ with quotes, i.e., \verb|<<|$sub ~ pred ~ obj$\verb|>>|, and $qt_{e}^{obj}$.obj means the object of the $qt_{e}^{obj}$.

\begin{figure}[!t]
\vspace{-1.0em}
\begin{algorithm}[H]
    \caption{QT-walk generation}
    \label{algorithm:gt_walk_generation}
    \begin{algorithmic}[1]
    \small
    \REQUIRE $wl$, $walk$, $triples_{e}^{subj}$, $qt_{e}^{obj}$, $qt_{e}^{subj}$, $\alpha$, $\beta$, $\gamma$, $\delta$
    \ENSURE $wl$
    \STATE $rand_{oq}, rand_{qs}, rand_{qo}, rand_{sq}$ $\gets$ random numbers
    \STATE $newWalk$ $\gets$ a copy of $walk$
    \STATE $modes$ $\gets$ empty list
    \IF{$qt_{e}^{subj}$ is not none AND $rand_{qs} < \alpha$}
        \STATE append ``qt2subject'' to $modes$
    \ENDIF
    \IF{$qt_{e}^{obj}$ is not none AND $rand_{oq} < \beta$}
        \STATE append ``object2qt'' to $modes$
    \ENDIF
    \IF{$qt_{e}^{obj}$ is not none AND $rand_{qo} < \gamma$}
        \STATE append ``qt2object'' to $modes$
    \ENDIF
    \IF{$qt_{e}^{subj}$ is not none AND $rand_{sq} < \delta$}
        \STATE append ``subject2qt'' to $modes$
    \ENDIF
    \STATE $mode$ $\gets$ random($modes$)
    \IF{$mode$ is empty}
        \FOR{$triple$ $\in$ $triples_{e}^{subj}$}
            \STATE append predicate of $triple$ to $newWalk$
            \STATE append object of $triple$ to $newWalk$
            \STATE append $newWalk$ to $wl$
        \ENDFOR
    \ELSE
        \IF{$mode$ is ``qt2subject''}
            \IF{$walk$ is empty}
                \STATE append $qt_{e}^{subj}$.qt to $newWalk$
            \ENDIF
            \STATE append $qt_{e}^{subj}$.subj to $newWalk$
            \STATE append $qt_{e}^{subj}$.pred to $newWalk$
            \STATE append $qt_{e}^{subj}$.obj to $newWalk$
            \STATE append $newWalk$ to $wl$
        \ELSIF{$mode$ is ``object2qt''}
            \STATE $qt$ $\gets$ $qt_{e}^{obj}$.qt
            \IF{walk is empty}
                \STATE append $qt_{e}^{obj}$.obj to $newWalk$
            \ENDIF
            \STATE append $qt$ to $newWalk$
        \ELSIF{$mode$ is ``qt2object''}
            \IF{$walk$ is empty}
                \STATE append $qt_{e}^{obj}$.qt to $newWalk$
            \ENDIF
            \STATE append $qt_{e}^{obj}$.obj to $newWalk$
        \ELSIF{$mode$ is ``subject2qt''}
            \STATE $qt$ $\gets$ $qt_{e}^{subj}$.qt
            \IF{walk is empty}
                \STATE append $qt_{e}^{subj}$.subj to $newWalk$
            \ENDIF
            \STATE append $qt$ to $newWalk$
        \ENDIF
        \STATE remove $walk$ from $wl$
        \STATE append $newWalk$ to $wl$
    \ENDIF
    \end{algorithmic}
\end{algorithm}
\end{figure}

\section{Relationship between graph walk depth and LP accuracy}
\label{appendix:depth_ref}

\begin{figure}[h!]
\begin{center}
\includegraphics[width=\linewidth]{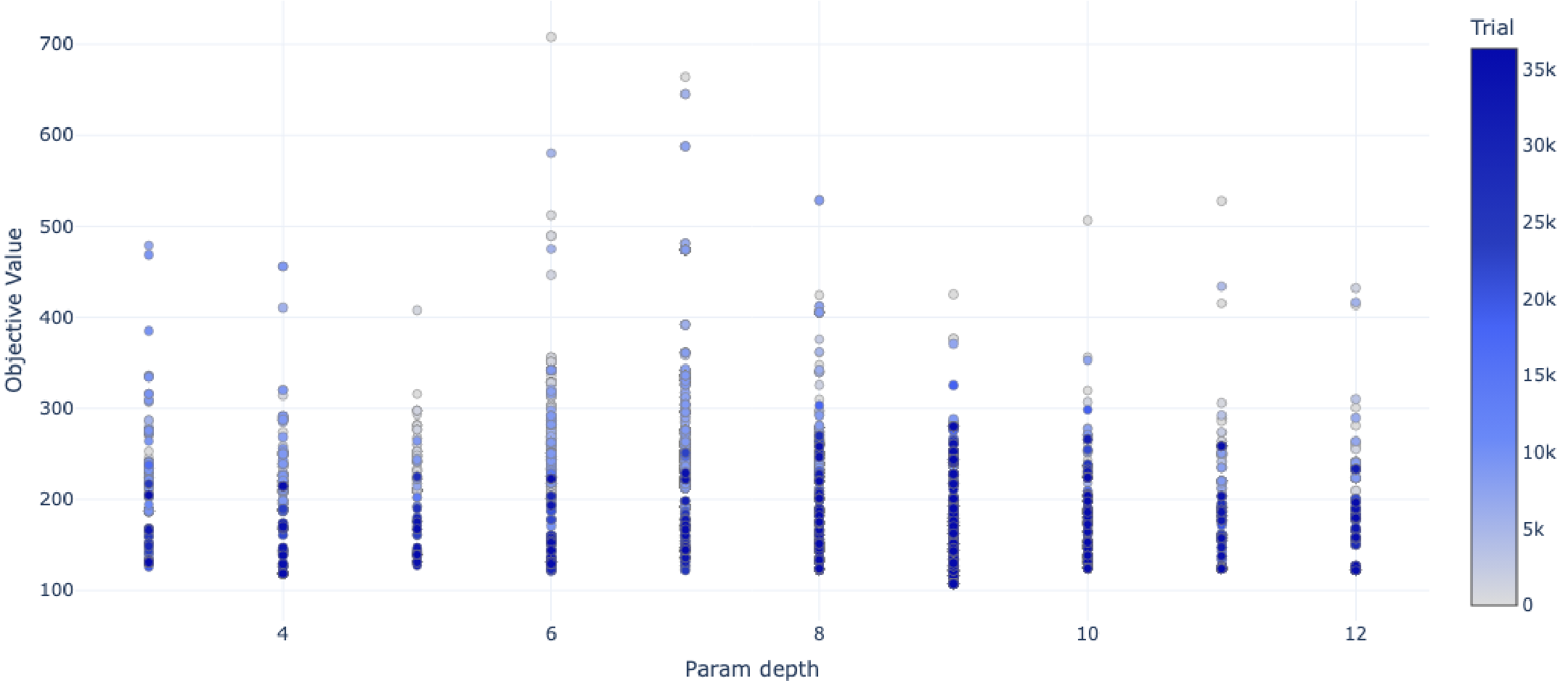}
\caption{Relationship between graph walk depth and LP accuracy (REF)}
\label{fig:depth_ref}
\end{center}
\end{figure}

\begin{figure}[h!]
\begin{center}
\includegraphics[width=\linewidth]{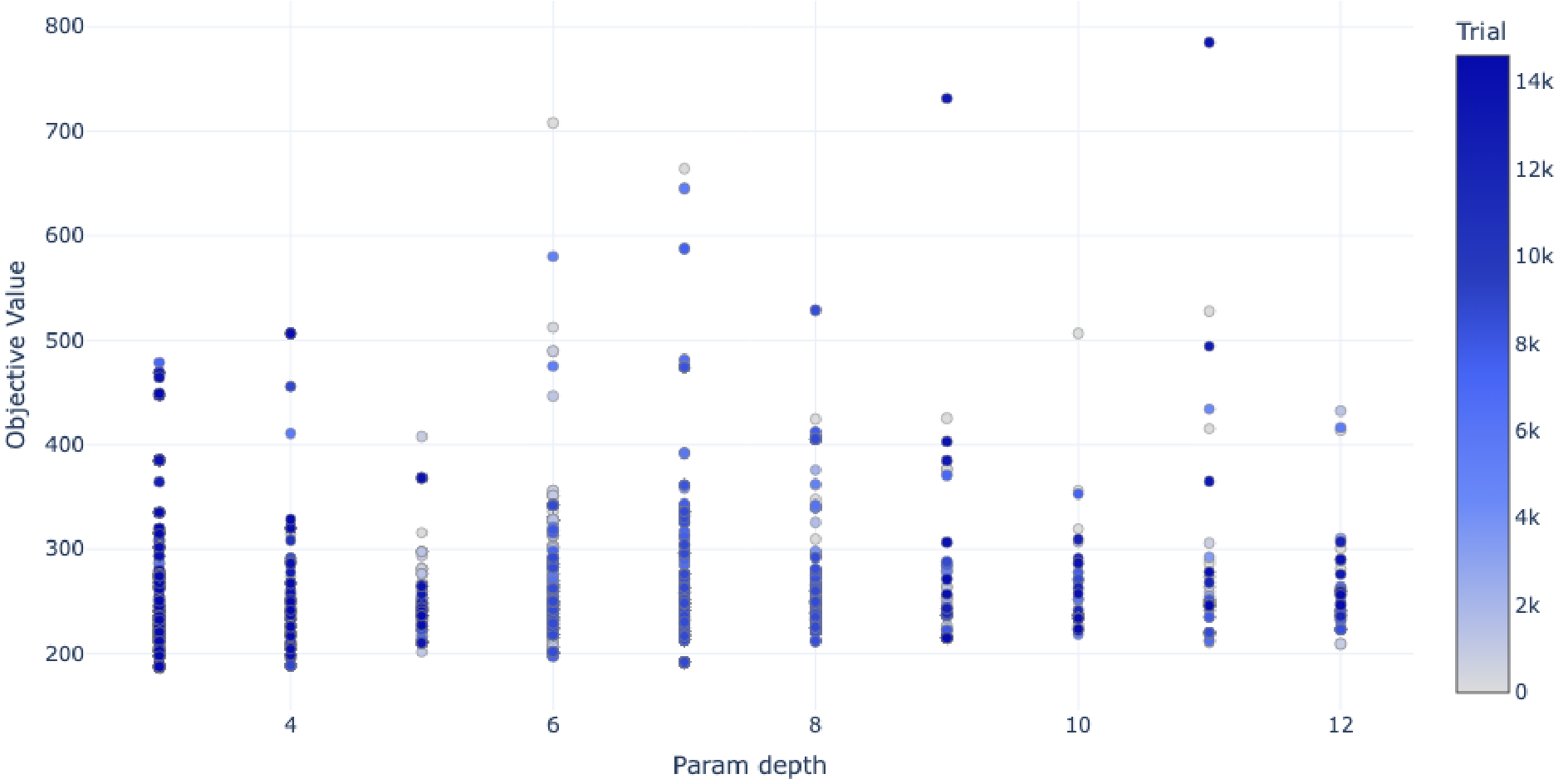}
\caption{Relationship between graph walk depth and LP accuracy (SGP)}
\label{fig:depth_sgp}
\end{center}
\end{figure}

\begin{figure}[h!]
\begin{center}
\includegraphics[width=\linewidth]{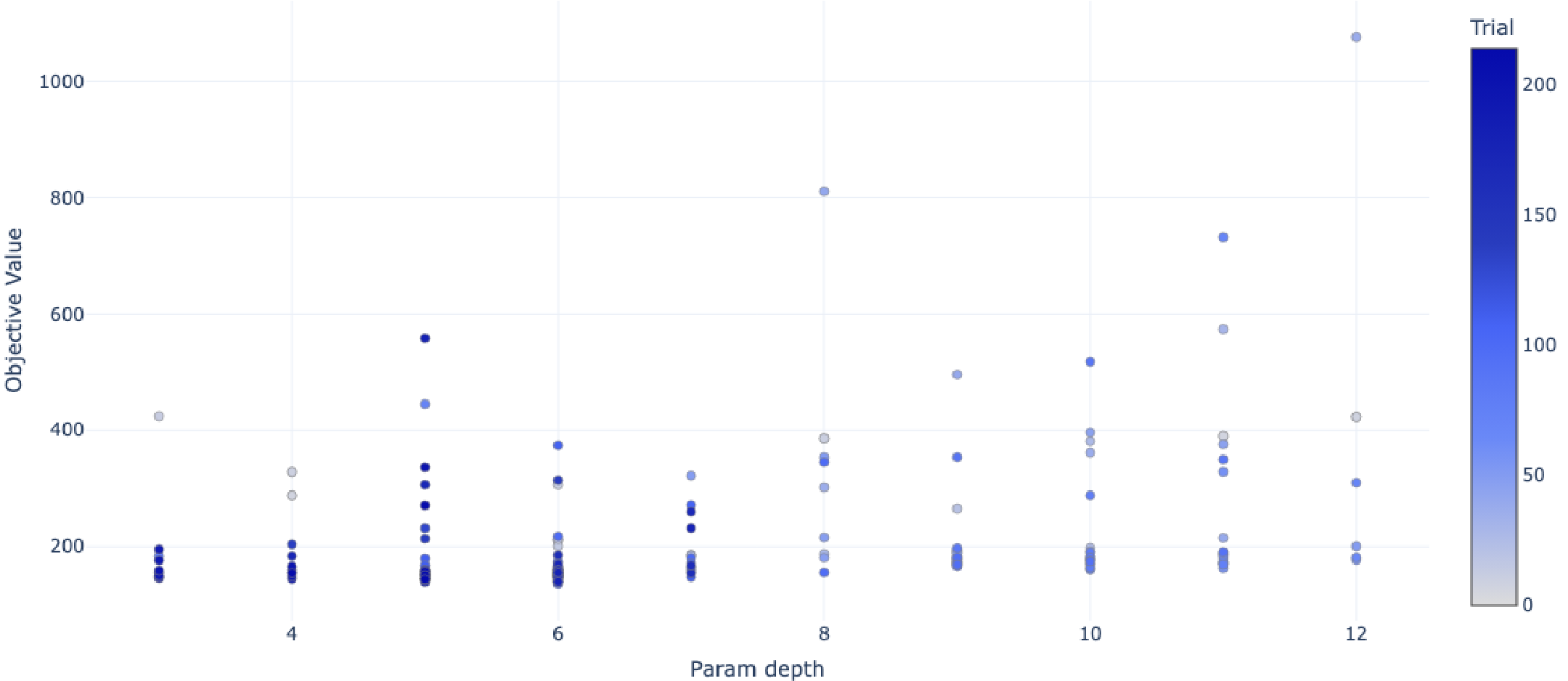}
\caption{Relationship between graph walk depth and LP accuracy (RDR)}
\label{fig:depth_rdr}
\end{center}
\end{figure}

In REF and SGP, the walk depth parameter of RDF-star2Vec has the most significant impact on LP accuracy. Figure~\ref{fig:depth_ref}, \ref{fig:depth_sgp}, and \ref{fig:depth_rdr} show the relationships between walk depth and LP accuracy on the KGRC-RDF dataset.

\section{Relationship between QT-walk transition probability and LP accuracy}
\label{appendix:qtwalk}

\begin{figure}[h!]
\begin{center}
\includegraphics[width=\linewidth]{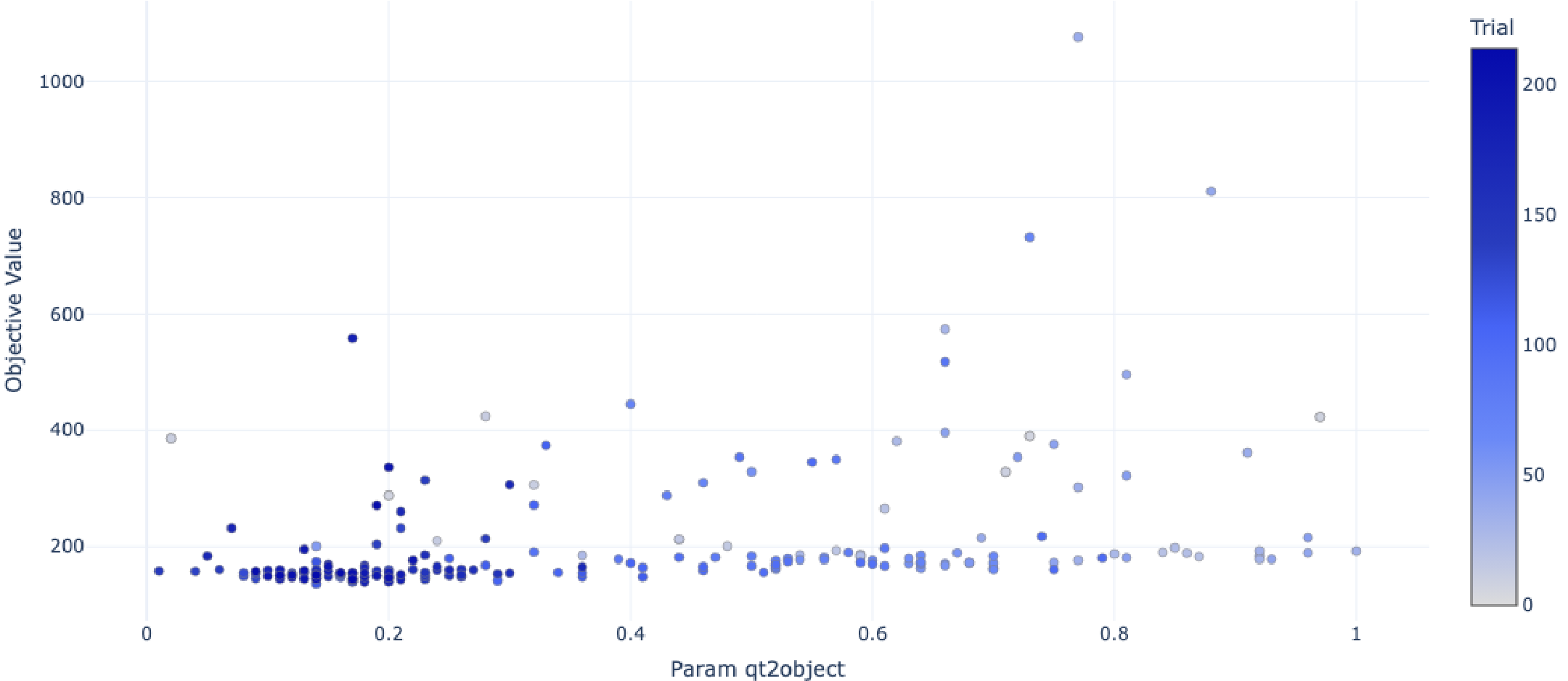}
\caption{Relationship between qo-walk (qt2object) transition probability and LP accuracy}
\label{fig:qt2object}
\end{center}
\end{figure}

\begin{figure}[h!]
\begin{center}
\includegraphics[width=\linewidth]{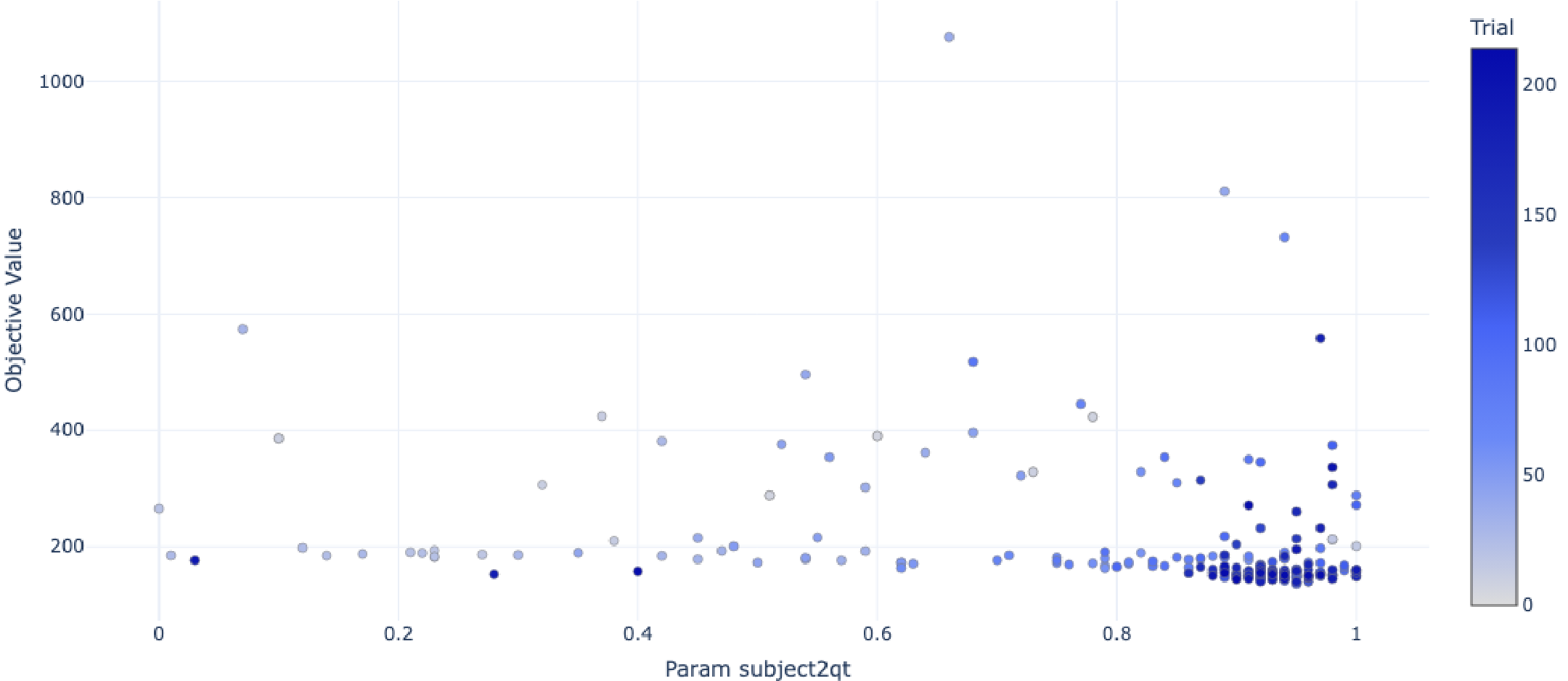}
\caption{Relationship between sq-walk (subject2qt) transition probability and LP accuracy}
\label{fig:subject2qt}
\end{center}
\end{figure}

In RDR, the transition probabilities of qo-walk (qt2object) and sq-walk (subject2qt) have a more significant impact on LP accuracy than the depth of the walk.
Figure~\ref{fig:qt2object} and \ref{fig:subject2qt} show the relationships between the transition probabilities of qo-walk and sq-walk and LP accuracy on the KGRC-RDF dataset.

\bibliographystyle{elsarticle-num}
\bibliography{ref}




%% For numbered reference style

% \end{thebibliography}
\end{document}